# Feature Selection via Probabilistic Outputs

**Nicholas A. Arnosti**                                    NARNOSTI@STANFORD.EDU

314X Huang Engineering Center, 475 Via Ortega, Stanford, CA 94305

**Andrea Pohoreckyj Danyluk**                             ANDREA@CS.WILLIAMS.EDU

Williams College, 47 Lab Campus Drive, Williamstown, MA 01267 USA

## Abstract

This paper investigates two feature-scoring criteria that make use of estimated class probabilities: one method proposed by Shen et al. (2008) and a complementary approach proposed below. We develop a theoretical framework to analyze each criterion and show that both estimate the spread (across all values of a given feature) of the probability that an example belongs to the positive class. Based on our analysis, we predict when each scoring technique will be advantageous over the other and give empirical results validating our predictions.

## 1. Introduction

Data sets used to perform classification often contain redundant and/or irrelevant information. Eliminating unhelpful features can reduce the computational complexity of many learning algorithms, increase the interpretability of the models they produce, and decrease the risk of over-fitting. For these reasons, a great deal of work has been dedicated to the task of feature selection (Guyon & Elisseeff, 2003).

This paper explores probabilistic feature selection techniques. We present a feature-scoring criterion based on the work of Shen et al. (2008) and develop a novel theoretical framework to analyze their score and ours. We show that their score approximates an upper-bound for the improvement in accuracy that each feature offers to the Bayes-optimal classifier. We demonstrate that both their scoring method and ours estimate the spread (across all values of a given feature) of the probability that an example $\mathbf{x}$ belongs to the positive class. Finally, we begin to characterize when each scoring technique proves advantageous over the other.

In the next section we introduce notation, and follow with a discussion of feature selection methods. In Section 4 we present the scoring criteria and our theoretical analysis. Section 5 outlines our predictions for the relative performance of the scores and gives preliminary empirical results. We close by discussing directions for future work.

## 2. Notation

Here we introduce notation that will be used in the remainder of this paper. We say that a training set consists of examples of the form $(\mathbf{x}_i, y_i)$. Each $\mathbf{x}_i$ is a vector of feature values taken from the space $\mathcal{X}$, and $y_i$ is a class label taken from the set $\mathcal{Y}$. The goal is to find a function $g : \mathcal{X} \to \mathcal{Y}$ such that for unseen examples $(\mathbf{x}, y) \in \mathcal{X} \times \mathcal{Y}$, $g(\mathbf{x}) = y$. We use $n$ to refer to the number of training examples and $d$ to represent the number of real-valued features for each example.

We use $p(\mathbf{x})$ to represent the density of the distribution from which feature values are drawn. Given a vector $\mathbf{x} \in \mathbb{R}^d$, $x^j$ refers to the value of the $j^{th}$ element of this vector and $\mathbf{x}^{-j} \in \mathbb{R}^{d-1}$ is the vector $\mathbf{x}$ with the $j^{th}$ feature removed. Thus we can equivalently express the density $p(\mathbf{x})$ as $p(x^j|\mathbf{x}^{-j})p(\mathbf{x}^{-j})$. The notation $\mathbb{P}(y = 1|\mathbf{x})$ represents the true probability that the example $\mathbf{x}$ belongs to class 1, and $\hat{\mathbb{P}}(y = 1|\mathbf{x})$ is an estimate of this probability. For binary classification, we take $\mathcal{Y} = \{-1, +1\}$.

Though all of our notation assumes real-valued features, the analysis presented applies equally well to variables that take on discrete values.

## 3. Feature Selection

Feature selection approaches generally fall into three broad categories: *filter*, *wrapper*, and *embedded* methods. Filters are effectively pre-processing steps that score features according to some criterion and select those with the highest scores. They are fast and simple, but tend to perform less well than other approaches, partly because they measure the impact of a feature without taking into account the way the classifier will use that feature. Embedded methods involve modification to the training algorithm itself so that features can be selected as part of the training process. They are usually classifier-specific, and thus less general than other selection techniques. For more on these approaches, see Guyon & Elisseeff (2003).

In this paper we focus on wrapper methods, which score a set of features according to the loss (on a test set) of a classifier trained using only these features (Guyon & Elisseeff, 2003). This approach considers variables in the context of others, can apply to virtually any classifier, and explicitly measures the use of the feature to the chosen classification algorithm.

A commonly-implemented wrapper approach, known as *recursive feature elimination*, greedily constructs nested subsets of features. Starting with the full set of features, a series of classifiers are trained. The variables that cause performance to suffer least when not used are eliminated, and the process repeats.

The most common criticism of wrapper methods is that naive implementations tend to be quite slow. Even using the greedy method described above, for a data set with $d$ features, $\mathcal{O}(d^2)$ classifiers must be trained. This can be prohibitively expensive, so a variety of methods have been developed to approximate the result of this process (Guyon et al., 2002; Maldonado & Weber, 2009).

## 4. Feature Selection Using Probabilistic Outputs

In classification tasks, the most commonly used loss function is simply the number of classification errors on the test set, or some close variant such as $F$-measure. This approach can have difficulty identifying significant features in high-dimensional spaces, where the influence of each feature tends to be small and removing any single feature is unlikely to notably affect classification performance.

One way to address this concern is to use algorithms that output estimated class probabilities. Since a feature may influence probability estimates without changing the predicted class label, scoring features according to their effect on probability estimates is more sensitive than considering only the misclassification rate.

In this section we examine a feature-scoring method that incorporates class probabilities proposed by Shen et al. (2008), and introduce our modified feature-scoring criterion. Both scores are intended to be used as part of a recursive feature elimination scheme. For ease of exposition, this section assumes a binary classification problem.

### 4.1. Two Feature Scoring Criteria

As discussed above, the sensitivity of class probabilities provides a natural measure of each feature's importance. Shen et al. (2008) propose the following feature ranking criterion based on this idea.

$$S_S(j) = \int_{\mathcal{X}} |\mathbb{P}(y = 1|\mathbf{x}) - \mathbb{P}(y = 1|\mathbf{x}^{-j})| p(\mathbf{x}) d\mathbf{x}. \quad (1)$$

Because it is impossible to measure the above quantities, the joint density $p(\mathbf{x})$ and the probabilities $\mathbb{P}(y = 1|\mathbf{x})$ and $\mathbb{P}(y = 1|\mathbf{x}^{-j})$ must be estimated. Shen et al. propose four techniques to approximate (1). They report the best results when using the following estimate:

$$\hat{S}_S(j) = \frac{1}{n} \sum_{i=1}^{n} |\hat{\mathbb{P}}(y = 1|\mathbf{x}_i) - \hat{\mathbb{P}}(y = 1|\mathbf{x}_i^{-j})|. \quad (2)$$

The motivation for the scoring system implied by $\hat{S}_S$ is clear: features that cause significant changes in the estimated class probabilities are ranked higher than those that do not. At the same time, it seems that an ideal importance measure should take into account not only the magnitude of the change in probability estimates, but also its sign.

To illustrate this point, suppose that we have the following classification task with two binary attributes:

| $x^1$ | $x^2$ | $p(\mathbf{x})$ | $\mathbb{P}(y=1|\mathbf{x})$ | $\mathbb{P}(y=1|x^1)$ | $\mathbb{P}(y=1|x^2)$ |
|---|---|---|---|---|---|
| 0 | 0 | 10/22 | 0.495 | 0.45 | 0.66 |
| 0 | 1 | 1/22 | 0.000 | 0.45 | 0.00 |
| 1 | 0 | 10/22 | 0.825 | 0.75 | 0.66 |
| 1 | 1 | 1/22 | 0.000 | 0.75 | 0.00 |

Note that for an example with features $\mathbf{x} = (0, 0)$, $\mathbb{P}(y = 1|\mathbf{x}) = 0.495$, whereas if information about $x^1$ is not available, we see that $\mathbb{P}(y = 1|\mathbf{x}^{-1}) = 0.66$. Though including the first feature in our model changes our estimated probabilities by 0.165 for *all* examples with feature vector $(0, 0)$, this change is only *beneficial* 50.5% of the time, since in the other 49.5% of cases, the example belongs to the positive class. Equation (2) does not take this fact into account.

In light of this observation, we propose the following scoring criterion:

$$\hat{S}_A(j) = \frac{1}{n} \sum_{i=1}^{n} y_i(\hat{\mathbb{P}}(y=1|\mathbf{x}_i) - \hat{\mathbb{P}}(y=1|\mathbf{x}_i^{-j})) \quad (3)$$

This score rewards features when their inclusion moves estimated probabilities towards the correct class and punishes for examples such that including the feature worsens our prediction.

We now have two proposed feature scoring functions given by $\hat{S}_S$ and $\hat{S}_A$. Our next goal is to develop a theoretical framework to assist in analyzing and understanding these measures.

### 4.2. Analysis of the score $S_S$

In this section, we analyze the quantity $S_S$ given in (1). This analysis is motivated by the thought that before dedicating too much effort to approximating $S_S$, we would like to ensure that it is a reasonable surrogate for the importance of a feature. The contributions of this section are two-fold. First, we demonstrate that $S_S$ provides an upper-bound for the improvement in accuracy exhibited by the Bayes-optimal classifier due to the inclusion of the $j^{th}$ feature. We also show that $S_S$ measures the expected mean absolute deviation of $\mathbb{P}(y=1|\mathbf{x})$ as the $j^{th}$ feature varies.

Ideally, $S_S$ should correspond in some way to the utility of the $j^{th}$ feature. Of course, the utility of a feature depends on the procedure by which the data are used (i.e., the classification algorithm of choice). Though Shen et al.'s work focuses on support vector machines (SVMs), we proceed with a general analysis in this section. Rather than measuring the utility of a feature to any particular classifier, we consider the utility provided by that feature to the best *possible* classifier.

If the true function $\mathbb{P}(y=1|\mathbf{x})$ were known, prediction error could be minimized by always predicting the most likely class. This decision rule defines the Bayes-optimal classifier, and for binary classification its expected accuracy is given by

$$\int_{\mathcal{X}} \max\{\mathbb{P}(y=1|\mathbf{x}), 1 - \mathbb{P}(y=1|\mathbf{x})\} p(\mathbf{x})d\mathbf{x}. \quad (4)$$

It is natural to measure each feature's importance by the improvement in accuracy that it offers the Bayes-optimal classifier:

$$S_B(j) = \int_{\mathcal{X}} \max\{\mathbb{P}(y=1|\mathbf{x}), 1 - \mathbb{P}(y=1|\mathbf{x})\} p(\mathbf{x})d\mathbf{x}$$

$$- \int_{\mathcal{X}} \max\{\mathbb{P}(y=1|\mathbf{x}^{-j}), 1 - \mathbb{P}(y=1|\mathbf{x}^{-j})\} p(\mathbf{x})d\mathbf{x}, \quad (5)$$

which can be re-expressed using the identity $\max\{p, 1-p\} = 1/2 + |p - 1/2|$ as

$$S_B(j) = \quad (6)$$

$$\int_{\mathcal{X}} \left( |\mathbb{P}(y=1|\mathbf{x}) - 1/2| - |\mathbb{P}(y=1|\mathbf{x}^{-j}) - 1/2| \right) p(\mathbf{x})d\mathbf{x}$$

The reverse triangle inequality implies that

$$S_B(j) \leq \quad (7)$$

$$\int_{\mathcal{X}} |(\mathbb{P}(y=1|\mathbf{x}) - 1/2) - (\mathbb{P}(y=1|\mathbf{x}^{-j}) - 1/2)| p(\mathbf{x})d\mathbf{x}$$

$$= \int_{\mathcal{X}} |\mathbb{P}(y=1|\mathbf{x}) - \mathbb{P}(y=1|\mathbf{x}^{-j})| p(\mathbf{x})d\mathbf{x} = S_S(j).$$

Thus, $S_S(j)$ provides an upper-bound for $S_B(j)$ that holds regardless of the number of features, the shape of their joint density, or the form of $\mathbb{P}(y=1|\mathbf{x})$.

The following computations serve to provide additional intuition for the quantity measured by $S_S(j)$. We can rewrite $S_S(j)$ as:

$$\int_{\mathbf{x}^{-j}} \left( \int_{\mathcal{X}^j} |\mathbb{P}(y=1|\mathbf{x}) - \mathbb{P}(y=1|\mathbf{x}^{-j})| p(x^j|\mathbf{x}^{-j})dx^j \right) p(\mathbf{x}^{-j})d\mathbf{x}^{-j} \quad (8)$$

The quantity inside of parentheses is the expected value of $|\mathbb{P}(y=1|\mathbf{x}) - \mathbb{P}(y=1|\mathbf{x}^{-j})|$ given $\mathbf{x}^{-j}$. Because $\mathbb{P}(y=1|\mathbf{x}^{-j}) = \mathbb{E}[\mathbb{P}(y=1|\mathbf{x})|\mathbf{x}^{-j}]$, this equals the mean absolute deviation (MAD) of $\mathbb{P}(y=1|\mathbf{x})$ given $\mathbf{x}^{-j}$ (where $MAD[Z] = \mathbb{E}|Z - \mathbb{E}[Z]|$). Thus,

$$S_S(j) = \mathbb{E} \, MAD[\mathbb{P}(y=1|\mathbf{x})|\mathbf{x}^{-j}]. \quad (9)$$

A visualization of this fact is presented in Figure 1. Intuitively, the most important features are those that cause significant fluctuation in $\mathbb{P}(y=1|\mathbf{x})$, so using a measure of the spread of $\mathbb{P}(y=1|\mathbf{x})$ as a feature ranking criterion seems reasonable. For entirely irrelevant features, $\mathbb{P}(y=1|\mathbf{x})$ does not vary as $x^j$ does, so $MAD[\mathbb{P}(y=1|\mathbf{x})|\mathbf{x}^{-j}] = 0$.

### 4.3. Analysis of Our Alternative Score

Having provided two reasons that $S_S$ seems like a reasonable feature scoring criterion, we now analyze our modified score $\hat{S}_A$ given in (3).

We first show that using the score $\hat{S}_A(j)$ as the criterion for feature removal is equivalent to performing recursive feature elimination using absolute loss, i.e. $L(\hat{\mathbb{P}}(y=1|\mathbf{x}_i), y_i) = |t_i - \hat{\mathbb{P}}(y=1|\mathbf{x}_i)|$, where $t_i$ is 1 if $y_i = 1$ and 0 otherwise. To see this, note that a recursive feature elimination algorithm using this loss function removes the feature $j$ that minimizes

$$\frac{1}{n}\sum_{i=1}^{n}|t_i - \hat{\mathbb{P}}(y=1|\mathbf{x}_i^{-j})| = \frac{1}{n}\sum_{i=1}^{n}y_i(t_i - \hat{\mathbb{P}}(y=1|\mathbf{x}_i^{-j}))$$
$$= \frac{1}{n}\sum_{i=1}^{n}y_i(t_i - \hat{\mathbb{P}}(y=1|\mathbf{x}_i)) + \hat{S}_A(j). \tag{10}$$

Since the first of these two terms has no $j$ dependence, it follows that the two algorithms yield identical feature rankings. By contrast, using the score $\hat{S}_S(j)$ does not correspond to any natural loss function.

We now establish that $\hat{S}_A(j)$ estimates a quantity proportional to the expected variance of $\mathbb{P}(y=1|\mathbf{x})$ given $\mathbf{x}^{-j}$. Note that an example with feature values $\mathbf{x}$ contributes $\hat{\mathbb{P}}(y=1|\mathbf{x}) - \hat{\mathbb{P}}(y=1|\mathbf{x}^{-j})$ to the sum in (3) if it belongs to the positive class, and the negative of this quantity otherwise. Since $\mathbb{E}[y_i|\mathbf{x}_i] = 2\mathbb{P}(y=1|\mathbf{x}) - 1$, given an example $\mathbf{x}$, its expected contribution to (3) is

$$(2\mathbb{P}(y=1|\mathbf{x}) - 1)(\hat{\mathbb{P}}(y=1|\mathbf{x}) - \hat{\mathbb{P}}(y=1|\mathbf{x}^{-j})). \tag{11}$$

Weighting the space $\mathcal{X}$ with its density function, it follows that the quantity approximated by $\hat{S}_A$ is:

$$S_A(j) = \tag{12}$$
$$\int_{\mathcal{X}}(2\mathbb{P}(y=1|\mathbf{x}) - 1)(\mathbb{P}(y=1|\mathbf{x}) - \mathbb{P}(y=1|\mathbf{x}^{-j}))p(\mathbf{x})d\mathbf{x}$$

Since $\mathbb{E}[\mathbb{P}(y=1|\mathbf{x})] = \mathbb{P}(y=1) = \mathbb{E}[\mathbb{P}(y=1|\mathbf{x}^{-j})]$, the piece of the integrand corresponding to the -1 vanishes, leaving

$$S_A(j) = 2\int_{\mathcal{X}}\mathbb{P}(y=1|\mathbf{x})^2 - \mathbb{P}(y=1|\mathbf{x})\mathbb{P}(y=1|\mathbf{x}^{-j})p(\mathbf{x})d\mathbf{x}. \tag{13}$$

As in (8), we move the integral with respect to the $j^{th}$ feature inside and observe that

$$\mathbb{E}[\mathbb{P}(y=1|\mathbf{x})\mathbb{P}(y=1|\mathbf{x}^{-j})|\mathbf{x}^{-j}] = \mathbb{P}(y=1|\mathbf{x}^{-j})^2, \tag{14}$$
$$= \mathbb{E}[\mathbb{P}(y=1|\mathbf{x})|\mathbf{x}^{-j}]^2.$$

Using the fact that $\mathrm{Var}[Z] = E[Z^2] - E[Z]^2$, we get

$$S_A(j) = 2\mathbb{E}[\mathrm{Var}(\mathbb{P}(y=1|\mathbf{x}))|\mathbf{x}^{-j}]. \tag{15}$$

This holds regardless of the dimension $d$, the density $p(\mathbf{x})$, or form of $\mathbb{P}(y=1|\mathbf{x})$. Because the variance measures the spread of a distribution, this indicates that $S_A$ (like $S_S$) matches our intuitive understanding that features to which $\mathbb{P}(y=1|\mathbf{x})$ is very sensitive are more important.

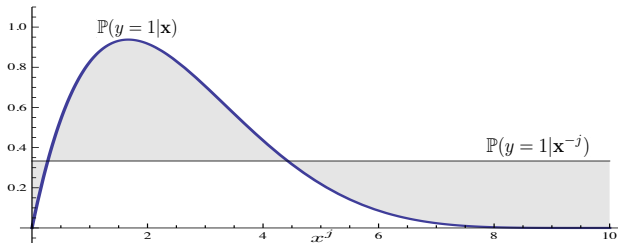

*Figure 1.* A hypothetical dependence of $\mathbb{P}(y=1|\mathbf{x})$ on $x^j$, for a fixed choice of $\mathbf{x}^{-j}$. If $p(x^j|\mathbf{x}^{-j})$ is uniform on $[0, 10]$, the gray region represents $MAD[\mathbb{P}(y=1|\mathbf{x})|\mathbf{x}^{-j}]$. $S_S(j)$ is the expected area of the gray region across all possibilities for $\mathbf{x}^{-j}$.

We have established that both $S_S(j)$ and $S_A(j)$ measure the expected variation in $\mathbb{P}(y=1|\mathbf{x})$ along the $j^{th}$ dimension. Intuitively, this suggests that $S_S(j)$ and $S_A(j)$ should be "similar." In fact, we can show that the following chain of inequalities holds:

$$0 \leq S_A(j), S_B(j) \leq S_S(j) \leq \sqrt{S_A(j)/2} \leq 1/2. \tag{16}$$

To help visualize the above inequalities, $S_S$, $S_A$, and $S_B$ are plotted in Figure 2 for a variation of the synthetic problem from Weston & Watkins (1999) [1].

It is worth noting that although $S_S$ and $S_A$ are closely related, this does not mean that they rank features identically. In the task presented in Section 4.1, $S_S(x^1) = 0.15 > S_S(x^2) = 0.10\overline{9}$ and $S_A(x^1) = 0.0495 < S_A(x^2) = 0.0765$, so even though true probabilities are known, $S_S$ and $S_A$ select different features.

## 5. Experimental Evaluation

In practice, $S_S$ and $S_A$ cannot be directly computed and must be estimated by $\hat{S}_S$ and $\hat{S}_A$, respectively. In this section we make predictions regarding $\hat{S}_S$ and $\hat{S}_A$ and provide results from several tests comparing them.

Computing either $\hat{S}_S$ or $\hat{S}_A$ is essentially a two-step process. First, the training data is used to fit the func-

---

[1] In Weston's problem, most features are noise following a normal distribution, while the informative features are all of the form:

$$X_{c,p} \sim \begin{cases} yN(c, 1) & : \text{with probability } p \\ N(0, 1) & : \text{otherwise} \end{cases}$$

Thus, even informative features take noisy values with probability $1 - p$. Note that features of this form are increasingly informative as $c$ and $p$ increase. This method for generating features has since been adopted elsewhere, including by Shen et al. (2008).

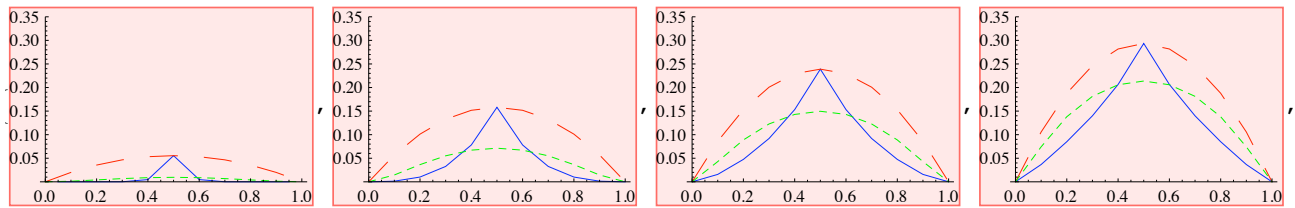

*Figure 2.* Scores assigned by $S_S$ (red, large dashes), $S_A$ (green, small dashes), and $S_B$ (blue, solid) to $X_{c,0.7}$ in isolation for $c \in (0.2, 0.6, 1.0, 1.4)$. The $x$-axis represents the percentage of examples belonging to the positive class. When classes are balanced, $S_S$ and $S_B$ are equal, and $S_S$ is always at least as large as the other two.

tion $\hat{\mathbb{P}}(y = 1|\mathbf{x})$. Then, this estimate is used to compute the sums in (2) and (3), which approximate the density $p(\mathbf{x})$. It came to our attention that because $\hat{S}_S$ does not use the labels of each example in this second step, the sum in (2) can be taken over any mix of labeled and unlabeled examples. In many domains, large sets of unlabeled examples are available while labeled training data is relatively sparse, so using unlabeled data in this way could substantially improve the performance of $\hat{S}_S$.

Both $\hat{S}_S$ and $\hat{S}_A$ approximate the term $\mathbb{P}(y = 1|\mathbf{x}) - \mathbb{P}(y = 1|\mathbf{x}^{-j})$ by assuming a functional form for $\mathbb{P}(y = 1|\mathbf{x})$ and fitting parameter values to the training data. Even with a large training set, these estimates may be significantly biased if $\mathbb{P}(y = 1|\mathbf{x})$ does not have the assumed form. We argue that including $y_i$ in $\hat{S}_A$ may alleviate this problem. To illustrate our point, suppose that $\mathbf{x}_0$ is such that $\mathbb{P}(y = 1|\mathbf{x}_0) = 1/2$. Then even if the estimates $\hat{\mathbb{P}}(y = 1|\mathbf{x}_0)$ and $\hat{\mathbb{P}}(y = 1|\mathbf{x}_0^{-j})$ are terribly wrong, as the number of training examples with feature vector $\mathbf{x}_0$ grows, we see a law of large numbers effect: the 50% of positive examples with feature vector $\mathbf{x}_0$ should cancel the corresponding negative examples in the sum from (3). By contrast, bad estimates for $\hat{\mathbb{P}}(y = 1|\mathbf{x}_0)$ and $\hat{\mathbb{P}}(y = 1|\mathbf{x}_0^{-j})$ could cause a large contribution when computing $\hat{S}_S(j)$ via (2), regardless of the training set size.

Our work from Section 4 and the above discussion lead us to the following predictions:

- Because variance is the square of standard deviation, which in turn is closely related to the mean absolute deviation, the score from $\hat{S}_A$ should vary approximately as the square of $\hat{S}_S$.
- When the training set is sampled disproportionately from the domain, applying $\hat{S}_S$ to large sets of unlabeled examples will provide a more consistent estimate of each feature's utility than either $\hat{S}_A$ or $\hat{S}_S$ applied only to the labeled training data.
- In cases where the assumed model does not fit the true dependence between $\mathcal{X}$ and $\mathcal{Y}$, $\hat{S}_A$ should identify relevant variables more reliably than $\hat{S}_S$.

We test the first two predictions in Sections 5.1 and 5.2, while leaving the third for future work. Additionally, we present preliminary results from applying $\hat{S}_S$ and $\hat{S}_A$ to a real-world data set in Section 5.3.

### 5.1. Relationship Between $\hat{S}_S$ and $\hat{S}_A$: a Simple Case

To confirm the prediction that $\hat{S}_A$ should vary approximately as the square of $\hat{S}_S$, we ran a trial on the synthetic variables $X_{c,p}$ as described earlier, with $c$ ranging from 0 to 3 in steps of 0.25 and $p$ in $[0, 1]$ with steps of 0.1. For each $(c, p)$ pair, we trained a linear SVM on a two-variable training set, where the first variable was $X_{c,p}$ and the second was Gaussian noise.

Support vector machines typically do not provide probability estimates, but Vapnik (1998), Hastie & Tibshirani (1998) and Platt (1999) have all proposed methods for using SVMs to generate probabilities. For tests in this section, we obtained and modified source code from Shen et al., which derives probability estimates using the technique proposed in (Platt, 1999) and an SVM implementation from LIBSVM. Results from these trials are shown in Figure 3, and confirm our prediction beautifully.

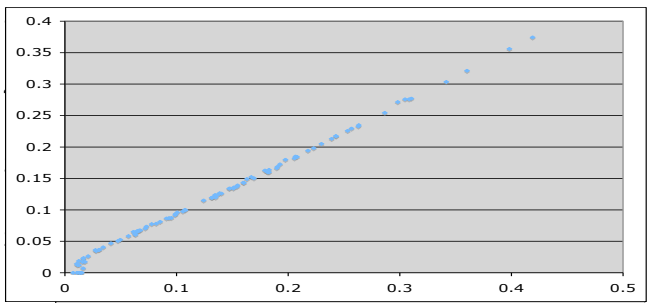

*Figure 3.* $\hat{S}_S(X_{c,p})$ ($x$-axis) plotted against $\sqrt{\hat{S}_A(X_{c,p})/2}$, with each point representing a different $(c, p)$ pair. Though these values are computed using (2) and (3) respectively, $\hat{S}_S \approx \sqrt{\hat{S}_A/2}$, just as predicted by (1) and (12).

## 5.2. Incorporating Unlabeled Examples

We designed the next test to explore the potential benefit of using unlabeled examples when computing $\hat{S}_S$. In this section, we use $\hat{S}_{SU}$ to represent the score (2) augmented by additional unlabeled examples and $\hat{S}_S$ to stand for the same score applied only to the labeled training instances.

For the purposes of this experiment, we constructed a synthetic data set with five nominal features. Each feature is independent of all others and takes on three possible values, say a, b, and c. The frequency with which each variable takes each value is given below:

|   | $x^1$ | $x^2$ | $x^3$ | $x^4$ | $x^5$ |
|---|---|---|---|---|---|
| a | 1/4 | 1/4 | 1/3 | 1/3 | 1/2 |
| b | 1/2 | 1/2 | 1/3 | 1/6 | 1/4 |
| c | 1/4 | 1/4 | 1/3 | 1/2 | 1/4 |

All but the first feature are noise (i.e. they do not affect the probability that the example belongs to the positive class). We chose $\mathbb{P}(y = 1 | x^1 = \mathtt{a}) = 1/4$, $\mathbb{P}(y = 1 | x^1 = \mathtt{b}) = 1/2$, $\mathbb{P}(y = 1 | x^1 = \mathtt{c}) = 3/4$.

Tests described in this section and 5.3 were conducted with a Naive Bayes classifier designed for nominal features. We made this choice because Naive Bayes classifiers explicitly provide estimated class probabilities. Additionally, they afford a natural (and efficient) way to compute $\hat{\mathbb{P}}(y = 1 | \mathbf{x}^{-j})$: missing values are handled by simply not including probabilities from that feature (Kononenko, 1991). Note that on this data set the features are independent, so with adequate training data a Naive Bayes classifier should replicate the true class probabilities.

Experiments were conducted in three trials. Each trial contained tests on nine different sizes of training sets. For each training size and each trial, 500 runs were performed. A single run for a specified trial and training size consisted of sampling an appropriate number of training examples, training a Naive Bayes classifier on these examples, and using this classifier (along with a collection of unlabeled examples when appropriate) to compute $\hat{S}_A(j)$, $\hat{S}_S(j)$, and $\hat{S}_{SU}(j)$ for $j = 1 \dots 5$. For each algorithm, we recorded the number of times (out of 500 runs) that it successfully ranked the first feature as the most informative.

The difference between the three trials was the manner in which training examples were selected. During the first trial, examples were drawn from the original distribution. During the second, the $x^1$ training values were equally likely to be a, b, and c (effectively undersampling from b and oversampling from a and c, thereby making $x^1$ seem more informative than it actually is). In the third trial, $x^1$ took the values a and c each with probability 1/8, and took the value b with probability 3/4. This has the effect of overstating the importance of $x^1$. In all trials, the dependency $\mathbb{P}(y = 1 | \mathbf{x})$ remained unchanged and the unlabeled examples provided to $\hat{S}_S$ were drawn from the original distribution. This design allowed us to explore the question of how each scoring system fares when there is bias in the process of selecting training examples. The results, shown for different training set sizes, are displayed in Figure 4.

When sampling proportionately from the data set, the three scores performed comparably. On the second trial (which we refer to as the "oversampling" trial due to the fact that it overstates the importance of $X^1$), $\hat{S}_S$ and $\hat{S}_A$ identified $x^1$ as the top feature more often than in the first trial (as expected), while $\hat{S}_{SU}$ performed nearly identically to the first trial. This suggests that our estimate for $\mathbb{P}(y = 1 | \mathbf{x})$ is sufficiently accurate that incorporating unlabeled examples yields scores as if the training examples had been sampled proportionately. On the third ("undersampling") trial, both $\hat{S}_S$ and $\hat{S}_A$ identified $x^1$ as the most important feature in fewer cases than either of the other trials. The degradation is most notable for $\hat{S}_S$. Meanwhile, $\hat{S}_{SU}$ performed at approximately the same level as on the other trials. The amount by which $\hat{S}_{SU}$ outperforms the other methods on the third trial appears to be independent of the number of training examples. These results suggest that when the training data is not representative of the entire domain and unlabeled examples are available, using them in scoring can be very beneficial. When no unlabeled examples are available, $\hat{S}_A$ provides a more reliable measure of each feature's importance than $\hat{S}_S$.

## 5.3. Breast Cancer Results

Here we test the performance of $\hat{S}_S$ and $\hat{S}_A$ on a real-world data set: Breast Cancer, available from the UCI repository (`http://archive.ics.uci.edu/ml/datasets/`). The task is to predict, based on 9 discrete features (each taking values in the set $\{1, 2, \dots, 10\}$), whether a tumor is malignant or benign.

The distribution of feature values is far from uniform: averaged across all features, 46.2% of values are 1, while only 1.1% take the value 9, and the values 6,7,8 each occur with frequency below 4%. Because we wished to see how $\hat{S}_S$ and $\hat{S}_A$ performed when presented with limited training data (in particular, as few as ten labeled examples), this meant that for any given training set, it was likely that most of the possible feature values would not be present in the training data.

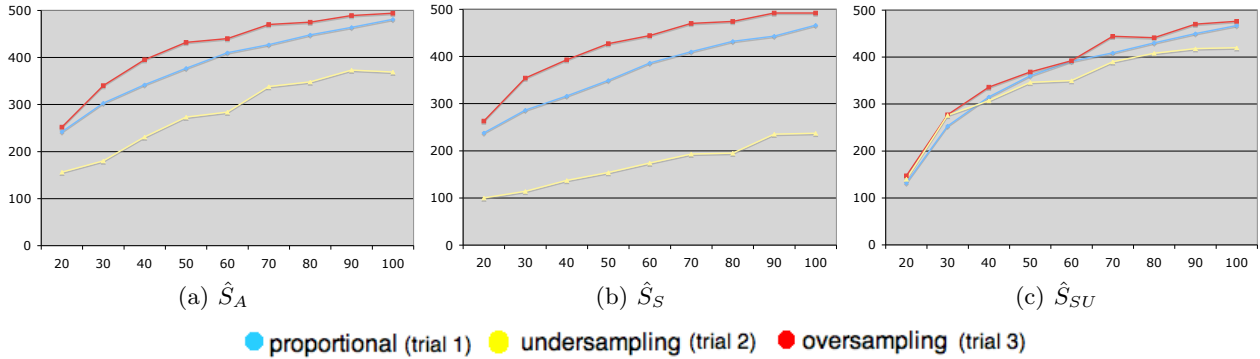

*Figure 4.* Number of times (out of 500) correctly picking the one informative feature from five total features, plotted against training set size. Between the two methods that use only labeled examples, $\hat{S}_A$ is less sensitive to changes in how training points are sampled than $\hat{S}_S$ is. Both methods suffer when the training data is undersampled from informative regions of the input space. Results for $\hat{S}_{SU}$ indicate that incorporating unlabeled examples provides a measure of each feature's utility that is far more robust to variations in the sampling of the training set.

To alleviate this problem, we converted each feature into a binary attribute according to the mapping $f$ given by:

$$f(\mathbf{x}) = \big(I(x^1 \leq 5), I(x^2 = 1), I(x^3 = 1), I(x^4 = 1),$$
$$I(x^5 \leq 2), I(x^6 = 1), I(x^7 \leq 3), I(x^8 = 1), I(x^9 = 1)\big).$$

Our basic classifier achieved a leave-one-out error rate of $23/699$ on the original data set. On the transformed data with binary features, this rate was $24/699$, indicating that for a Naive Bayes classifier, effectively no predictive power is lost by our transformation.

In order to easily evaluate $\hat{S}_S$ and $\hat{S}_A$, we augmented the data set by adding three purely noisy binary features. We conducted experiments on sets of size $10, 20, \ldots, 100$. For each size, 20 runs were conducted. A run consisted of sampling a set of positive and negative examples with replacement from the full data set.

Given a training set of $n$ examples, $n$ classifiers were trained, each using all but one example to compute estimates for $\hat{\mathbb{P}}(y = 1|\mathbf{x})$ and $\hat{\mathbb{P}}(y = 1|\mathbf{x}^{-j})$ on the held-out data point. These estimates were used to compute $\hat{S}_S$ and $\hat{S}_A$, thereby obtaining a ranking of the features.

For each training size, we computed the *aggregate rank* for each feature by ranking them on the basis of their average rank across all 20 runs. As shown in Table 1, for all training sizes and both scores, the three uninformative features were among the five features with the lowest aggregate rank. Additionally, across all runs and training sizes, neither score ever ranked one of our dummy features as its top choice.

When provided with at least 30 training examples, $\hat{S}_S$

ranked all three noisy features among the bottom four on all 20 runs. $\hat{S}_A$ did not perform quite as well: even with 100 training examples, on two of 20 runs one of the uninformative features was ranked as highly as fifth.

It is difficult to infer much from the ranking of the original features, because all of them are at least weakly predictive of the class label. $\hat{S}_S$ and $\hat{S}_A$ generally agreed that feature 9 was the least useful of the original features. As one might hope, when we tested the performance of classifiers trained on each possible pairs of features, those using $x^9$ performed least well.

## 6. Conclusions

In this paper, we have considered the feature scoring algorithm presented by Shen et al. (2008) and proposed our own related score for use in feature selection tasks. We focused on these techniques because they consider the importance of each variable in the context of others and can score variables even in high-dimensional contexts where each feature's impact on the final prediction is small. Our primary contribution is a careful analysis of Shen et al. (2008)'s score, $S_S$, and our alternative criterion $S_A$.

We demonstrated that the quantity $S_S(j)$ is the expected conditional mean absolute deviation of $\mathbb{P}(y = 1|\mathbf{x})$ given $\mathbf{x}^{-j}$, and that $S_A(j)$ is the expected conditional variance of $\mathbb{P}(y = 1|\mathbf{x})$ given $\mathbf{x}^{-j}$. These proofs suggest that each score is a reasonable criterion for feature selection, as $S_S$ and $S_A$ select as important the features whose values have the greatest influence on $\mathbb{P}(y = 1|\mathbf{x})$.

| $\hat{S}_S$ | 10 | 20 | 30 | 40 | 50 | 60 | 70 | 80 | 90 | 100 | $\hat{S}_A$ | 10 | 20 | 30 | 40 | 50 | 60 | 70 | 80 | 90 | 100 |
|---|---|---|---|---|---|---|---|---|---|---|---|---|---|---|---|---|---|---|---|---|---|
| $x^{10}$ | 9 | 11 | 12 | 12 | 12 | 12 | 12 | 12 | 12 | 12 | $x^{10}$ | 10 | 9 | 10 | 10 | 10 | 9 | 8 | 11 | 9 | 10 |
| $x^{11}$ | 11 | 12 | 11 | 11 | 11 | 11 | 11 | 11 | 11 | 11 | $x^{11}$ | 12 | 10 | 9 | 12 | 9 | 11 | 11 | 12 | 10 | 9 |
| $x^{12}$ | 10 | 9 | 10 | 10 | 10 | 10 | 10 | 10 | 10 | 10 | $x^{12}$ | 8 | 11 | 8 | 10 | 12 | 8 | 12 | 9 | 8 | 11 |

*Table 1.* Aggregate rank (among 12 features) given by $\hat{S}_S$ (left) and $\hat{S}_A$ (right) to the three noise features $x^{10}, x^{11}, x^{12}$ in our Breast Cancer Data set, for training sizes ranging from 10 to 100. Even when provided with only ten training examples, the aggregate rank for each noise feature was always among the bottom five. On individual runs, these features were occasionally ranked higher than $8^{th}$, though for $\hat{S}_S$ this occurred only when training with 10 or 20 examples.

As alternative justification for $S_S$ we proved that $S_S(j)$ provides an upper-bound for the improvement in accuracy of the Bayes-optimal classifier due to the information provided by the $j^{th}$ feature. Additionally, we hypothesized that the approximation $\hat{S}_S$ could benefit from unlabeled examples. For a simple synthetic task, we confirmed that this data improved the robustness of $\hat{S}_S$ to variations in the way that the labeled examples were sampled.

We motivated our score, $\hat{S}_A(j)$, by observing that it measures both the magnitude and the sign of changes in estimated class probabilities due to the $j^{th}$ feature. We proved that using $\hat{S}_A$ to eliminate features from the data set is equivalent to minimizing total loss on the training set when using an $L_1$ loss function. For the problem described in Section 5.2, we concluded that when there was no unlabeled data available to supplement the training set, $\hat{S}_A$ was less sensitive than $\hat{S}_S$ to sampling variations in the training data.

## 7. Future Work

We view this paper as a beginning, rather than conclusive, investigation of feature selection using probabilistic outputs. As such, there are many interesting directions for future work.

Much of the analysis here pertains to the quantities $S_S$ and $S_A$, but in practice we are forced to use approximations. An open question is the extent to which the approximations used in this paper are "good." One way to quantify this would be to give sufficient conditions for these estimates to converge to $S_S$ and $S_A$ as the number of training examples grows.

We argued in Section 5 that the fact that $\hat{S}_A$ incorporates the sign of changes in predictions should help to mitigate the presence of bias due to modeling assumptions. One major goal for the future is to validate this prediction, either empirically or theoretically.

The eventual goal of this work is to develop the theory of feature selection using probabilistic outputs to the point where, given a data set, we can choose a feature scoring algorithm that is likely to perform well in the specified domain. In order to accomplish this, we hope to perform tests on real-world data to determine the extent to which the theory developed in this paper extends to different learning algorithms and domains.